\documentclass[sigconf,screen]{acmart}

\AtBeginDocument{%
  }

\setcopyright{acmlicensed}
\copyrightyear{2025}
\acmYear{2025}
\acmDOI{XXXXXXX.XXXXXXX}
\acmConference[MM '25]    
{2025 ACM Multimedia Conference} 
{October 27--31, 2025}  
{Dublin, Ireland}      

\acmISBN{978-1-4503-XXXX-X/2018/06}

\usepackage{listings}
\usepackage{xcolor}

\begin{document}

\title{Beyond Vision: Contextually Enriched Image Captioning with Multi-Modal Retrieval}

\author{Nguyen Lam Phu Quy}
\authornote{The first three authors contributed equally as lead authors.}
\email{23122048@student.hcmus.edu.vn}
\affiliation{%
  \institution{University of Science - VNUHCM}
  \city{Ho Chi Minh City}
  \country{Vietnam}
}

\author{Pham Phu Hoa}
\authornotemark[1]
\email{23122030@student.hcmus.edu.vn}
\affiliation{%
  \institution{University of Science - VNUHCM}
  \country{Vietnam}
}

\author{Tran Chi Nguyen}
\authornotemark[1]
\email{23122044@student.hcmus.edu.vn}
\affiliation{%
  \institution{University of Science - VNUHCM}
  \city{Ho Chi Minh City}
  \country{Vietnam}
}

\author{Dao Sy Duy Minh}
\authornote{The last three authors contributed equally in a supporting role.}
\email{23122041@student.hcmus.edu.vn}
\affiliation{%
  \institution{University of Science - VNUHCM}
  \city{Ho Chi Minh City}
  \state{State B}
  \country{Vietnam}
}

\author{Nguyen Hoang Minh Ngoc}
\authornotemark[2]
\email{ng0005oc@e.ntu.edu.sg}
\affiliation{%
  \institution{Nanyang Technological University}
  \country{Singapore}
}

\author{Huynh Trung Kiet}
\authornotemark[2]
\email{23132039@student.hcmus.edu.vn}
\affiliation{%
  \institution{University of Science - VNUHCM}
  \city{Ho Chi Minh City}
  \country{Vietnam}
}

\begin{abstract}

 Real-world image captions often lack contextual depth, omitting crucial details such as event background, temporal cues, outcomes, and named entities that are not visually discernible. This gap limits the effectiveness of image understanding in domains like journalism, education, and digital archives, where richer, more informative descriptions are essential. To address this, we propose a multimodal pipeline that augments visual input with external textual knowledge. Our system retrieves semantically similar images using BEIT-3 (Flickr30k-384 and COCO-384) and SigLIP So-384, reranks them using ORB and SIFT for geometric alignment, and extracts contextual information from related articles via semantic search. A fine-tuned Qwen3 model with QLoRA then integrates this context with base captions generated by Instruct BLIP (Vicuna-7B) to produce event-enriched, context-aware descriptions. Evaluated on the OpenEvents v1 dataset, our approach generates significantly more informative captions compared to traditional methods, showing strong potential for real-world applications requiring deeper visual-textual understanding. Our code is available at \url{https://github.com/PhamPhuHoa-23/Event-Enriched-Image-Captioning-ReZeroSlavery}
\end{abstract}

\begin{teaserfigure}
  \includegraphics[width=\textwidth]{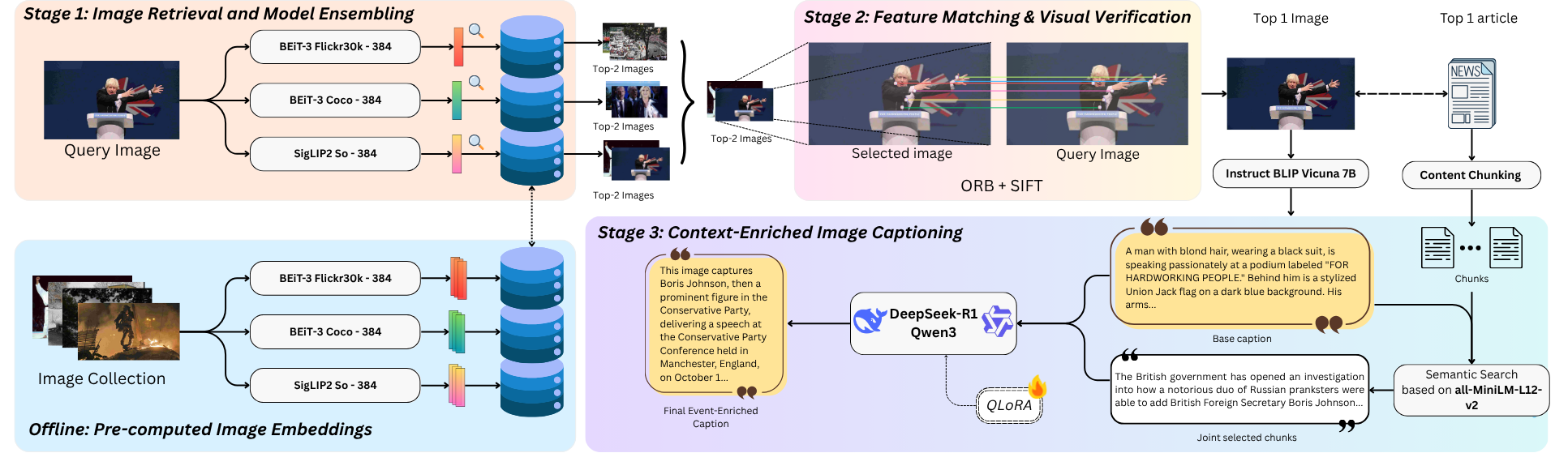}
  \caption{System Architecture of the Multimodal Caption Generation Framework}
  \Description{This is the system architecture of our pipeline}
  \label{fig:teaser}
\end{teaserfigure}

\received{1 July 2025}

\maketitle

\section{Introduction}

Image captioning has evolved from simple object recognition to sophisticated multimodal understanding, yet current approaches remain limited by their reliance on purely visual information \cite{li2023blip2bootstrappinglanguageimagepretraining, li2022blipbootstrappinglanguageimagepretraining}. While state-of-the-art models can accurately describe visible elements, they fail to capture crucial contextual details such as event background, temporal dynamics, named entities, and real-world significance that extend beyond what is directly observable in the image.

Several recent works have attempted to incorporate external knowledge into image captioning \cite{qu2024visuallyawarecontextmodelingnews, biten2019goodnewseveryonecontext, zhao2021boostingentityawareimagecaptioning}. However, these approaches typically generate brief, sentence-level descriptions that focus on immediate visual-textual alignment rather than comprehensive event understanding. For instance, knowledge-enhanced models may correctly identify "President Biden" in an image but fail to provide the rich contextual narrative about the specific meeting, its outcomes, or broader implications, producing captions of 10-20 words instead of the detailed, paragraph-length descriptions needed for meaningful event documentation.

This limitation is particularly problematic in domains like journalism and digital archives, where images serve as visual documentation of significant events requiring rich, contextually-aware descriptions. A photograph of officials at a conference table may appear visually similar across different contexts, but the underlying meaning—whether depicting peace negotiations, economic summits, or crisis meetings—fundamentally alters its significance and demands comprehensive explanation.

To address these challenges, we propose a novel retrieval-augmented captioning system that integrates external knowledge with visual analysis to generate detailed, paragraph-length captions. Our approach combines semantic image retrieval using BEIT-3\cite{beit3, beitv2, beit} and SigLIP\cite{zhai2023sigmoidlosslanguageimage} encoders, geometric reranking with ORB\cite{DBLP:journals/corr/abs-1710-02726} and SIFT \cite{DBLP:journals/corr/abs-1710-02726} features, contextual information extraction from related articles, and caption generation using a fine-tuned DeepSeek-R1-0528-Qwen3-8B\cite{deepseekai2025deepseekr1incentivizingreasoningcapability} model with QLoRA adapters \cite{dettmers2023qloraefficientfinetuningquantized}.

Our system generates captions that capture: (1) named entities and specific attributes beyond visual recognition, (2) temporal context and event dynamics, (3) coherent narratives connecting visual content to real-world significance, and (4) human-aligned writing style with comprehensive detail. Evaluated on the OpenEvents v1 dataset \cite{nguyen2025openeventsv1largescalebenchmark}, our approach demonstrates significant improvements over baseline methods, transforming image captioning from simple object enumeration into contextually-rich, detailed storytelling that bridges visual perception with comprehensive event understanding.

\section{Background and Related Work}
\subsection{OpenEvents v1}
OpenEvents v1 \cite{nguyen2025openeventsv1largescalebenchmark} is a large-scale, contextually rich, and challenging benchmark for event-enriched image captioning and event-driven image retrieval. This dataset is derived from a curated corpus of 202,803 news articles and 415,324 images from two major international news outlets: CNN and The Guardian.

Owing to its reflection of real-world image analysis, especially in the context of narrative news articles, this dataset demands models that can effectively bridge textual and visual modalities, encouraging deeper semantic and event-level comprehension. 

Accordingly, we leveraged this dataset to train and evaluate our model that bridges textual and visual content, fostering deeper semantic and event-level comprehension. 

\subsection{BEiT-3 and SigLIP2}
\subsubsection{BEiT-3}
BEiT-3 \cite{beit3} is a general-purpose multimodal foundation model designed to process images, text, and image-text pairs in a unified manner. It leverages a shared Transformer backbone with Multiway Transformers \cite{wang2022imageforeignlanguagebeit}, which includes lightweight modality-specific adapters to preserve modality-awareness while maintaining generalization. This architecture enables parameter efficiency and allows BEiT-3 to learn aligned representations across vision and language.

In our work, we use BEiT-3 to embed both articles and images into a shared semantic space. This unified representation facilitates tasks such as cross-modal retrieval and contextual image captioning, offering strong semantic alignment without requiring contrastive learning. Its versatility and strong pretraining make it well-suited for bridging textual and visual content in our multimodal pipeline.

\subsubsection{SigLIP 2}
SigLIP2 \cite{tschannen2025siglip2multilingualvisionlanguage} is a multilingual vision-language encoder that extends the original SigLIP \cite{zhai2023sigmoidlosslanguageimage} model by integrating multiple training strategies into a unified framework, including captioning-based pretraining, self-supervised losses (self-distillation, masked prediction), and online data curation. Unlike CLIP, which uses a contrastive objective, SigLIP2 employs a sigmoid-based binary classification loss for image-text alignment, allowing for more flexible learning of semantic relationships. It also incorporates a Transformer decoder during pretraining for caption generation and referring expression tasks, enhancing its localization and grounding capabilities.

\subsection{ORB and SIFT}
In this paper, we leveraged both ORB (Oriented FAST and Rotated BRIEF) \cite{6126544} and SIFT (Scale Invariant Feature Transform) \cite{article1} for feature matching, taking advantage of their complementary strengths to achieve robustness across diverse image types. In experiments involving various image transformations, including scaling, shearing, fish eye distortion, etc., ORB consistently demonstrated faster performance with lightweight computation, making it suitable for real-life applications. In contrast, SIFT delivered higher accuracy and precision, particularly in challenging scenarios. By combining both methods, our system benefits from the speed of ORB and the robustness of SIFT, ensuring reliable performance across a wide range of visual conditions of images in the dataset \cite{DBLP:journals/corr/abs-1710-02726}.

\subsection{InstructBLIP (Vicuna‑7B) (Salesforce/instructblip‑vicuna‑7b)}
InstructBLIP \cite{dai2023instructblipgeneralpurposevisionlanguagemodels} is a vision-language model designed to follow natural language instructions for a wide range of multimodal tasks, including image captioning, visual question answering (VQA), and image-based reasoning. It builds upon the BLIP-2 \cite{li2023blip2bootstrappinglanguageimagepretraining} architecture by introducing instruction tuning with over 15 million image-instruction pairs, significantly improving the model’s ability to generate grounded and contextually appropriate responses. The architecture integrates a frozen image encoder, a Q-former module, and a language model—in our case, Vicuna 7B.

Compared to earlier models like LLaVA \cite{liu2023visual}  and BLIP-2 \cite{li2022blipbootstrappinglanguageimagepretraining}, InstructBLIP demonstrates superior performance on tasks requiring detailed reasoning and descriptive generation, making it especially suitable for generating factual and coherent base captions. Its strong instruction-following behavior allows the use of carefully crafted prompts to elicit rich visual descriptions that align with the visual input, thus serving as an effective foundation for downstream caption refinement in context-enriched settings.
\subsection{Context Chunking and Embedding using all-MiniLM-l12-v2}
To identify contextually relevant text segments from associated articles, we employ \texttt{all-MiniLM-L12-v2} \cite{reimers2019sentence, wang2020minilm}, a lightweight yet high-performing sentence embedding model from the SentenceTransformers library. This model is based on the MiniLM architecture, which distills larger Transformer models (e.g., BERT \cite{devlin2019bertpretrainingdeepbidirectional}, RoBERTa \cite{liu2019robertarobustlyoptimizedbert}) into a smaller footprint while retaining strong semantic representation capabilities.

\texttt{all-MiniLM-L12-v2} encodes sentences into fixed-size dense vectors optimized for semantic similarity tasks using contrastive learning. It has been shown to perform competitively on the STS (Semantic Textual Similarity) and IR (Information Retrieval) benchmarks despite having only 33M parameters, making it ideal for fast, scalable retrieval in constrained-resource or multi-stage pipelines.

In our system, it serves as the embedding backbone for computing cosine similarity between the base caption (generated from visual input) and all sentence chunks extracted from news articles. Its efficiency enables rapid semantic filtering over large textual corpora, ensuring that only the most relevant and information-rich chunks are selected for downstream caption refinement.
\subsection{DeepSeek-R1-0528-Qwen3-8B}
For context-aware caption refinement, we employed the open-weight model \texttt{DeepSeek-R1-0528-Qwen3-8B-GGUF}, a chain-of-thought (CoT) distilled version of DeepSeek‑R1‑0528 \cite{deepseekai2025deepseekr1incentivizingreasoningcapability} built atop Qwen3‑8B \cite{yang2025qwen3technicalreport}. DeepSeek‑R1‑0528  is an enhanced reasoning model from DeepSeek that demonstrated state-of-the-art performance on math, programming, and logic benchmarks, rivaling larger models like OpenAI’s o1 \cite{openai2024openaio1card}. By distilling its CoT capabilities into the compact Qwen3‑8B base \cite{yang2025qwen3technicalreport}, this variant achieves exceptional reasoning depth—matching or surpassing Qwen3‑235B \cite{yang2025qwen3technicalreport} on AIME \cite{patel2024aimeaioptimizationmultiple} and related tasks—while maintaining an 8B parameter size.

\section{Methodology}
\subsection{Image Retrieval and Ensemble Models}
After obtaining embeddings from both BEiT-3 and SigLIP2, we perform model ensembling by combining their similarity scores, enabling more robust and comprehensive retrieval performance.

In this re-ranking phase, we created a unified candidate pool containing up to six elements, composed of the top two results from each of the three models. A weighting scheme is then applied as follows: 
\begin{itemize}
    \item Position weight: a score of 1.0 is assigned to top-1 results, and 0.8 to top-2 results. 
    \item Model weight: each model contributes equally with a weight of 1/n, where n is the number of models (in this case, n = 3).
\end{itemize}

The weighted score for each candidate is calculated as:
$$S_{weighted} = S_{original} \times W_{model} \times W_{position}$$

where $W_{model} = \frac{1}{n}$ (n = number of models) and $W_{position} = \{1.0, 0.8\}$ for top-1 and top-2 results respectively.

To further promote consensus among models, we apply a bonus of 
$0.03 \times (\text{number of appearances} - 1)$ to candidates that appear in multiple models' top results.

The final score incorporates consensus promotion:
$$S_{final} = \frac{1}{m}\sum_{i=1}^{m} S_{weighted,i} + 0.03 \times (m - 1)$$

where $m$ represents the number of models that retrieved the candidate in their top-2 results. Candidates are then sorted based on this final score to determine the most relevant matches.

The output from the re-ranking phase is then passed to the next stage, where image analysis is conducted through feature matching.

\subsection{Feature Matching \& Visual Verification}
The feature matching process begins by detecting keypoints using ORB and SIFT detectors, then comparing keypoints between query and candidate images. To ensure matching quality, we apply the Lowe's Ratio Test \cite{lowe2004distinctive} with threshold $\tau = 0.7$ to eliminate weak keypoint pairs, followed by RANSAC \cite{fischler1981random} to remove outliers.

We compute the inlier ratio as an indicator of visual similarity:
$$R_{inlier} = \frac{N_{inliers}}{N_{total\_matches}}$$

To evaluate spatial distribution, we divide the image into a 4×4 grid and compute spatial consistency:
$$S_{spatial} = \frac{N_{occupied\_cells}}{16} \times \exp\left(-\frac{\sigma_{grid}}{N_{inliers}}\right)$$

Scale consistency is measured by comparing keypoint size ratios:
$$S_{scale} = \frac{1}{1 + \sigma_{scale\_ratios}}$$

The final confidence score combines multiple factors:
\begin{itemize}
    \item 0.4 × inlier difference, 
    \item 0.3 × inlier ratio,
    \item 0.2 × homography score,
    \item 0.1 × spatial distribution consistency.
\end{itemize}

Reranking occurs when confidence exceeds 0.4 and the minimum inliers are greater than 8, ensuring robust visual verification beyond embedding similarity alone.
\subsection{Base Caption Generation}
To initiate the captioning pipeline, we utilized the InstructBLIP Vicuna 7B model \cite{dai2023instructblipgeneralpurposevisionlanguagemodels} to generate base captions that explicitly describe the visual content of each image. Rather than relying on generic prompts, we crafted a task-specific instruction after analyzing a wide range of sample captions in the training set. This analysis revealed recurring patterns and salient visual elements—such as people, objects, settings, activities, atmosphere, and spatial relationships—which are often critical for conveying event-relevant information. 

Guided by this insight, we designed the prompt to explicitly elicit these elements:

\begin{quote}
\texttt{Describe this image in detail. Focus on the visible people, objects, setting, activities, lighting, atmosphere, and any notable elements that would be important for news reporting. Focus on factual details and spatial relationships to create a rich visual narrative.}
\end{quote}

This prompt formulation is central to our approach: it maximizes the extraction of structured visual information from the image, forming a more complete and coherent base representation. By encouraging the model to articulate the visual scene in a factual, grounded manner, we aim to reduce ambiguity and improve the downstream performance of the language model in the enrichment phase. This is particularly important because large language models (LLMs) that operate solely on text often suffer from a "visual bottleneck"—they lack direct perception of the image and must rely entirely on the initial caption to "imagine" the scene. A detailed base caption thus serves as a crucial bridge between vision and language, enabling the LLM to reason more effectively about event context.

The caption generation was carried out using the following decoding configuration:

\begin{itemize}
\item \texttt{max\_length = 350}, \texttt{min\_length = 50}
\item \texttt{length\_penalty = 1.2}, \texttt{repetition\_penalty = 1.12}
\item \texttt{num\_beams = 5}, \texttt{early\_stopping = True}
\item \texttt{pad\_token\_id = processor.tokenizer.eos\_token\_id}
\end{itemize}

To enhance robustness, erroneous or low-quality outputs were reprocessed using InstructBLIP Vicuna 14B \cite{dai2023instructblipgeneralpurposevisionlanguagemodels} with a similar setup.

\subsection{QLoRA-based Fine-tuning of DeepSeek Qwen3 for Context-Enriched Image Captioning}

To enrich the base captions with contextual information, we first segmented each article into overlapping text chunks using a sliding window of three sentences with a one-sentence stride. After chunking, we employed the \texttt{all-MiniLM-L12-v2}  sentence embedding model \cite{reimers2019sentencebertsentenceembeddingsusing} to perform semantic similarity search, using the pre-generated base caption as the query. The top five most relevant chunks were selected based on cosine similarity. To preserve the broader discourse structure of the article, we also included the first three and the last two sentences, ensuring that both the lead-in and conclusion were retained. Named entities (e.g., people, locations, organizations) were extracted using regular expressions to enhance factual grounding. All selected segments were concatenated with a task-specific prompt to form the final input to the language model.

We fine-tuned the \texttt{unsloth/DeepSeek-R1-0528-Qwen3-8B-GGUF} \cite{ionescu2024nonlinearlandaudampingwave} \cite{Aizpurua_2025} model using QLoRA \cite{dettmers2023qloraefficientfinetuningquantized} (Quantized Low-Rank Adaptation), with a rank of 256 and 8-bit precision to reduce memory overhead while maintaining model performance. The model was trained for two epochs using the standard Causal Language Modeling (CLM) \cite{radford2019language} objective, which minimizes the next-token prediction loss via cross-entropy:

\[
\mathcal{L}_{\text{CLM}} = - \sum_{t=1}^{T} \log P(x_t \mid x_{<t}; \theta)
\]

where \( x_t \) is the token at position \( t \), \( x_{<t} \) is the sequence prefix, and \( \theta \) denotes the model parameters. This objective not only guides the model to generate fluent and coherent text but also plays a key role in improving downstream caption quality as measured by metrics such as CIDEr \cite{vedantam2015cider}. Since CIDEr \cite{vedantam2015cider} emphasizes content overlap and semantic relevance with human-written references, optimizing the model to predict the next informative token within a structured prompt encourages better alignment with key content phrases and information units typically favored by CIDEr scoring \cite{vedantam2015cider}. We observed that training with CLM on focused, high-relevance input improves both lexical diversity and semantic density, contributing to higher CIDEr scores \cite{vedantam2015cider} without the need for custom loss terms.

In a comparative setup, we fine-tuned the same model using a larger rank of 512, while omitting the semantic search stage and instead feeding the entire article as context. Despite the increase in trainable parameters, this configuration resulted in slower convergence: the loss decreased gradually and plateaued at a higher value. Furthermore, the model tended to underfit early, showing limited improvement across epochs.

We hypothesize that this behavior arises from the input being too lengthy and noisy, leading to diluted supervision signals during training. Without semantic filtering, the model must parse a large amount of potentially irrelevant text, making it harder to focus on the salient information needed to condition the caption. In contrast, our semantic search approach pre-selects the most contextually relevant chunks, providing a stronger alignment between input and target caption. This focused conditioning likely facilitates faster and more stable optimization, especially under limited fine-tuning steps.

\noindent \textbf{Prompt Design.} The prompt used during training was carefully engineered to simulate the behavior of a seasoned news caption writer. Key instructions included:

\begin{enumerate}
    \item The NEWS CONTEXT is MORE IMPORTANT than visual details.
    \item Start with “The image shows” but immediately connect to the news story.
    \item Use 70\% article information + 30\% visual description (ensure coherence between visual elements and article content).
    \item Emphasize WHO, WHAT, WHY, WHEN, and WHERE extracted from the article.
    \item Mention specific names, organizations, and events referenced in the article.
    \item Explain the significance and broader implications of the news.
    \item Only describe visual elements that are relevant to the news story.
    \item Produce a 300–350 word caption prioritizing factual and newsworthy content.
\end{enumerate}

This setup enables the model to generate captions that are context-rich, informative, and stylistically aligned with journalistic standards, even without the need for additional loss functions beyond standard language modeling.

\begin{figure}[h]
  \centering
  \includegraphics[width=1.0\linewidth]{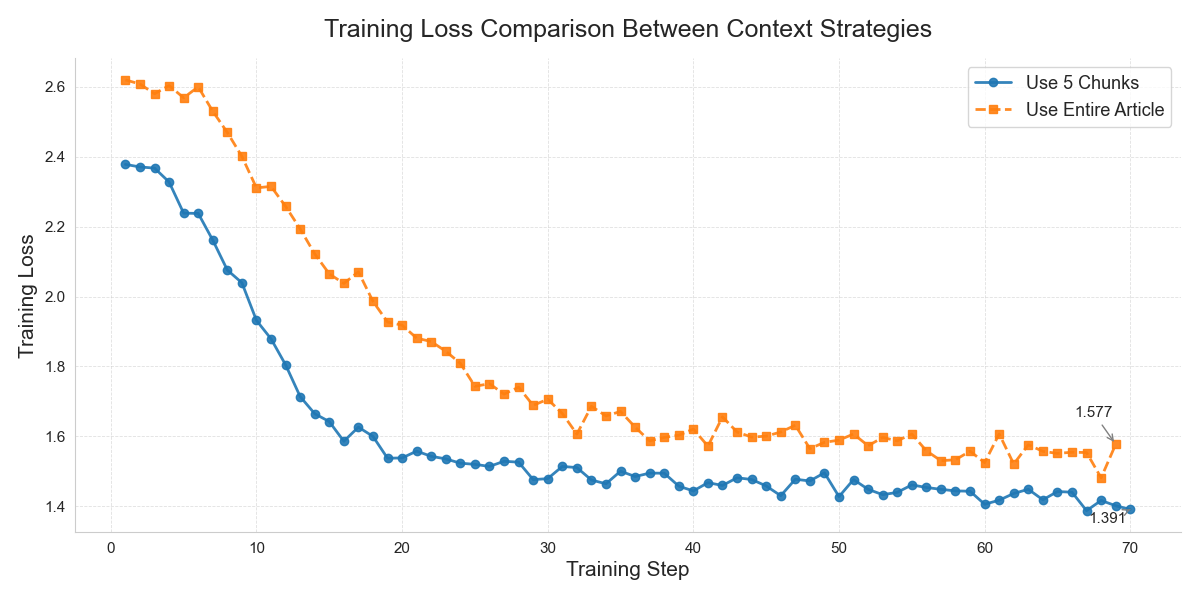}
  \caption{Training loss comparison: semantic chunk selection (rank 64) vs. full article input (rank 512).}
  \label{fig:loss_semantic_vs_full}
\end{figure}

\noindent \textbf{Comparison of Training Dynamics.} Figure~\ref{fig:loss_semantic_vs_full} illustrates the training loss curves for both configurations, highlighting the impact of semantic filtering on convergence speed and stability.

\section{Results}

Our system was evaluated on the EVENTA Grand Challenge CondaBench, achieving an overall score of \texttt{0.45148} on the hidden test set, which reflects performance across the following key metrics:

\begin{table}[H]
  \caption{Metrics of Private Test Set}
  \label{tab:results-hidden}
  \begin{tabular}{ccp{5cm}}
    \toprule
    Metric & Value & Interpretation \\
    \midrule
    \texttt{AP} & 0.955 & The average precision (AP) measures retrieval accuracy across different thresholds. A score of 0.955 demonstrates that our system maintains high precision throughout, reflecting robust and consistent retrieval quality. \\
    \texttt{Recall@1} & 0.945 & This indicates that in 94.5\% of cases, the correct item is ranked at the top of the retrieval results.  \\
    \texttt{Recall@10} & 0.973 & This shows that the correct item appears within the top 10 results in 97.3\% of the time, demonstrating strong retrieval breadth. \\
    \texttt{CLIPScore\cite{hessel2022clipscorereferencefreeevaluationmetric}} & 0.732 & This score reflects a high level of semantic alignment between the generated captions and the corresponding images. \\
    \texttt{CIDEr\cite{vedantam2015cider}} & 0.156 & The CIDEr score\cite{vedantam2015cider} measures the similarity between the generated captions and human-written references, with this value indicating moderate agreement in terms of content and phrasing. \\
  \bottomrule
\end{tabular}
\end{table}

We also report our model's performance on the public test set for comparability, following the official EVENTA evaluation protocol:

\begin{table}[H]
  \caption{Metrics of Public Test Set}
  \label{tab:results-public}
  \begin{tabular}{ccccccc}
    \toprule
    Overall & AP & Recall@1 & Recall@10 & CLIPScore & CIDEr \\
    \midrule
    \textbf{0.51815} & 0.994 & 0.990 & 1.000 & 0.748 & 0.195 \\
    \bottomrule
  \end{tabular}
\end{table}

Compared to existing baselines on the public test set (see Table~\ref{tab:baseline-public}), our system achieves significantly stronger performance across multiple dimensions. In particular:
\begin{itemize}
  \item Our CLIPScore\cite{hessel2022clipscorereferencefreeevaluationmetric}of \textbf{0.748} surpasses the best prior baseline (Gemma\cite{gemmateam2024gemmaopenmodelsbased} + Article, 0.6634) by \textbf{+12.8\%}, indicating a substantially better alignment between generated captions and images.
  \item Our CIDEr \cite{vedantam2015cider} of \textbf{0.195} is more than \textbf{10× higher} than most prior methods (e.g., Qwen\cite{bai2023qwentechnicalreport}: 0.0282, Gemma\cite{gemmateam2024gemmaopenmodelsbased}: 0.0111), demonstrating superior content fidelity to human references.
  \item We also outperform all evaluated vision-language (VL) baselines, including SmolVLM\cite{marafioti2025smolvlmredefiningsmallefficient}, Qwen\cite{bai2023qwentechnicalreport}, and Gemma\cite{gemmateam2024gemmaopenmodelsbased} (both with and without article context), across all key metrics.
\end{itemize}

These results highlight the effectiveness of our event-enriched captioning pipeline, which not only integrates contextual information effectively but also provides robust image-language grounding.

\begin{table}[H]
  \caption{Captioning results on the public test set (from baseline models)}
  \label{tab:baseline-public}
  \begin{tabular}{lcccc}
    \toprule
    Method & CLIPScore & CIDEr & BLEU-4 & METEOR \\
    \midrule
    SmolVLM & 0.4609 & 0.0044 & 0.0155 & 0.0789 \\
    SmolVLM + Article & 0.5552 & 0.0170 & 0.0229 & 0.0738 \\
    Qwen & 0.5283 & 0.0282 & 0.0256 & 0.1320 \\
    Qwen + Article & 0.5855 & 0.0565 & 0.0419 & 0.1383 \\
    Gemma & 0.5945 & 0.0111 & 0.0243 & 0.1322 \\
    Gemma + Article & 0.6634 & 0.0184 & 0.0341 & 0.1453 \\
    \bottomrule
  \end{tabular}
\end{table}

\section{Discussions and future directions}

While our current system performs well in both the retrieval and captioning phases, it still relies heavily on the DeepSeek-Qwen3\cite{deepseekai2025deepseekr1incentivizingreasoningcapability} model for generating the final captions. As a result, there is a risk of hallucinations, where the model may introduce details that are not grounded in the image or retrieved context. Moreover, the CIDEr score \cite{vedantam2015cider} remains relatively low, indicating that the generated captions, while informative, may not yet fully align with human-like phrasing or closely match reference captions in terms of fluency and specificity.

In future iterations, we aim to extend our architecture by replacing the DeepSeek-Qwen3 \cite{deepseekai2025deepseekr1incentivizingreasoningcapability}model with a range of large language models (LLMs), each trained on diverse domains, data distributions, and alignment strategies. By integrating diverse LLMs, we can explore how different model architectures and pretraining corpora influence the quality, specificity, and factuality of the generated captions. This comparative analysis will enable dynamic model selection for caption generation depending on the input image type, domain, or application context. 

In addition, we plan to experiment with alternative models for generating the initial (base) captions—especially \textbf{Qwen-VL 2.5  7B\cite{bai2025qwen25vltechnicalreport}}, a vision-language model with strong OCR capabilities, improved factual grounding, and entity recognition. Currently, we use Instruct BLIP-Vicuna-7B\cite{dai2023instructblipgeneralpurposevisionlanguagemodels} to generate base captions; however, its limited ability to extract named entities or interpret image-embedded text can result in vague or generic outputs. In contrast, Qwen-VL 2.5\cite{bai2025qwen25vltechnicalreport} exhibits superior capacity in recognizing scene-specific entities and reading text from the image, which we believe will substantially enhance the quality, richness, and factual accuracy of the base captions. These improved base captions will then serve as more informative inputs to the DeepSeek-Qwen3 \cite{deepseekai2025deepseekr1incentivizingreasoningcapability}model during the enrichment phase, ultimately leading to higher-quality final outputs and reducing the risk of hallucinated content.

We also intend to explore post-processing techniques to refine generated captions after decoding. These include re-ranking candidate outputs using retrieval-based scoring, fluency polishing via lightweight LLM-based editing, and applying rule-based or learned constraints to enforce relevance, conciseness, or stylistic consistency. Such strategies may help improve final metric scores (e.g., CIDEr \cite{vedantam2015cider}, METEOR\cite{robinson2015introducing}) and better align captions with human preferences.

\section{Conclusion}

In this work, we presented a novel multimodal pipeline for generating event-enriched image captions by combining visual similarity search, semantic content retrieval, and cross-modal fusion with a QLoRA\cite{dettmers2023qloraefficientfinetuningquantized} fine-tuned large language model. Evaluated on the OpenEvents v1 dataset\cite{nguyen2025openeventsv1largescalebenchmark}, our system demonstrates strong performance across a range of retrieval and captioning metrics, consistently outperforming traditional captioning approaches by incorporating external knowledge and domain-specific context.

This pipeline holds significant potential for real-world applications in domains such as news media, digital archives, and educational platforms, where interpretability, historical relevance, and semantic depth are essential.

\begin{acks}

We would like to express our sincere gratitude to the Software Engineering Laboratory (SELAB) at the University of Science, VNU-HCM, for organizing the meaningful EVENTA Grand Challenge as part of ACM Multimedia 2025. This competition provided us with a valuable opportunity to work with a rich multimodal dataset and develop models for event-enriched captioning.

\end{acks}

\bibliographystyle{ACM-Reference-Format}
\bibliography{sample-base}

\end{document}